# Artificial Neural Network based Diagnostic Model for Causes of Success and Failures

Bikrampal Kaur,
Deptt.of Computer Science & Engineering,
Chandigarh Engineering College,
Mohali,India.
dhaliwal_bikram@yahoo.com

Dr.Himanshu Aggrawal,
Deptt.of Computer Engineering,
Punjabi Unniversity,
Patiala,India.
himanshu@pbi.ac.in

*Abstract*— Resource management has always been an area of prime concern for the organizations. Out of all the resources human resource has been most difficult to plan, utilize and manage. Therefore, in the recent past there has been a lot of research thrust on the managing the human resource. Studies have revealed that even best of the Information Systems do fail due to neglect of the human resource. In this paper an attempt has been made to identify most important human resource factors and propose a diagnostic model based on the back-propagation and connectionist model approaches of artificial neural network (ANN). The focus of the study is on the mobile -communication industry of India. The ANN based approach is particularly important because conventional approaches (such as algorithmic) to the problem solving have their inherent disadvantages. The algorithmic approach is well-suited to the problems that are well-understood and known solution(s). On the other hand the ANNs have learning by example and processing capabilities similar to that of a human brain. ANN has been followed due to its inherent advantage over conversion algorithmic like approaches and having capabilities, training and human like intuitive decision making capabilities. Therefore, this ANN based approach is likely to help researchers and organizations to reach a better solution to the problem of managing the human resource. The study is particularly important as many studies have been carried in developed countries but there is a shortage of such studies in developing nations like India. Here, a model has been derived using connectionist-ANN approach and improved and verified via back-propagation algorithm. This suggested ANN based model can be used for testing the success and failure human factors in any of the communication Industry. Results have been obtained on the basis of connectionist model, which has been further refined by BPNN to an accuracy of 99.99%. Any company to predict failure due to HR factors can directly deploy this model.

*Keywords*— Neural Networks, Human resource factors, Company success and failure factors.

## I. INTRODUCTION

Achieving the information system success is a major issue for the business organizations. Prediction of a company's success or failure is largely dependent on the management of human resource (HR). Appropriate utilization of human resource may lead to the success of the company and their underutilization may lead to its failure.

In most of the organizations management makes use of conventional Information System (IS) for predicting the management of the human resources. In this paper an attempt have been made to identify and suggest HR factors and propose a model to determine the influence of HR factors leading to failure. It is particularly important as the neural networks have proved their potential in several fields such as Industry, transport, dairy sectors etc.. India has distinguished IT strength in global scenario and using technologies like neural networks is extremely important due to their decision making capabilities like human brain.

In this paper a Neuro-Computing approach has been proposed with some metrics collected through pre acquisition step from the communication industry. In this study, a coding of backpropagation algorithium have been used to predict success or failure of company and also a comparison is made with the connectionist model for predicting the results. The back-propagation learning algorithm based on gradient descent method with adaptive learning mechanism.. The configuration of the connectionist approach has also been designed empirically. To this effect, several architectural parameters such as data pre-processing, data partitioning scheme, number of hidden layers, number of neurons in each hidden layer, transfer functions, learning rate, epochs and error goal have been empirically explored to reach an optimum connectionist network.

## II. REVIEW OF LITERATURE

The review of IS literature suggests that for the past 15 years, the success and the failure HR factors in information systems have been major concern for the academics, practitioners, business consultants and research organizations.

A number of researchers and organizations throughout the world have been studying that why information systems do fail, some important IS failure factors identified by [6,7] are:

- Critical Fear-based culture.
- Technical fix sought.
- Poor reporting structures
- Poor consultation.
- Over commitment.
- Changing requirements.
- Political pressures.
- Weak procurement.

- Technology focused.
- Development sites split.
- Leading edge system
- Project timetable slippage
- Complexity underestimated
- Inadequate testing.
- Poor training

Six major dimensions of IS viz. superior quality (the measure of IT itself), information quality (the measure of information quality), information use (recipient consumption of IS output), user satisfaction (recipient response to use of IS output), individual impact (the impact of information on the behavior of the recipient) and organizational impact (the impact of information on organizational performance) had already been proposed [8] All these dimensions directly or indirectly are related to HR of IS.

Cancellation of IS projects [11] are usually due to a combination of:
- Poorly stated project goals
- Poor project team composition
- Lack of project management and control
- Little technical know-how
- Poor technology base or infrastructure
- Lack of senior management involvement
- Escalating project cost and time of completion

Some of the other elements of failure [12] identified were:
- Approaches to the conception of systems;
- IS development issues (e.g. user involvement)
- Systems planning
- Organizational roles of IS professionals
- Organizational politics
- Organizational culture
- Skill resources
- Development practices (e.g. participation)
- Management of change through IT
- Project management
- Monetary impact of failure
- "Soft" and Hard" perceptions of technology
- Systems accountability
- Project risk
- Prior experience with IT
- Prior experience with developing methods
- Faith in technology
- Skills, attitude to risk

All the studies predict that during the past two decades, investment in Information technology and Information system have increased significantly in the organization. But the rate of failure remains quite high. Therefore an attempt is made to prepare the HR model for the prediction of the success or failure of the organization.

### III. OBJECTIVES OF STUDY

(i) To design a HR model of factors affecting success and failure in Indian Organisations of Telecom sectors.
(ii) To propose a diagnostic ANN based model of the prevailing HR success/failure factors in these organizations.

A model depicting important human resources factors has been designed on the basis of literature survey and researchers experiences in the industry under this study has been in figure1.

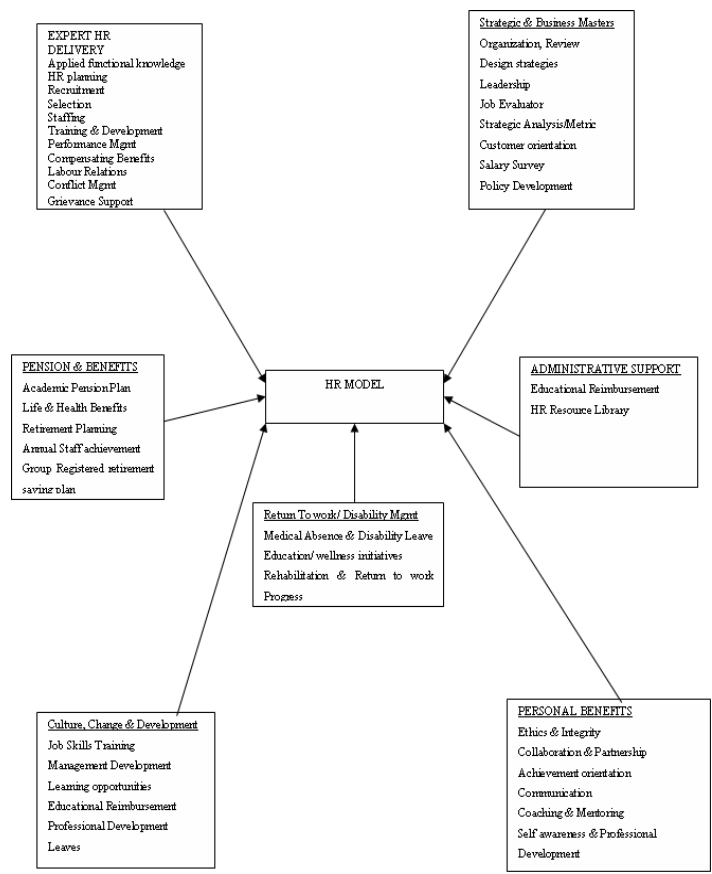

Fig.-1 Exhaustive View of HR Model

### IV. EXPERIMENTAL SCHEME

#### A. Using ANN

Neural networks differ from conventional approach of problem solving in a way similar to the human brain. The network is composed of a large number of highly interconnected processing elements (neurons) working in parallel to solve a specific problem. Neural networks learn by example. Differences in ANN and conventional systems have been given below in TABLE I.

TABLE-1
COMPARISON OF ANN AND CONVENTIONAL SYSTEM

| S.No | ANN | Conventional Systems |
|---|---|---|
| 1 | Learn by examples | Solve problems by algorithmic approach |
| 2 | Unpredictable | Highly predictable & well defined |
| 3 | Better decision making due to human like intelligence | No decision making |
| 4 | Trial and error method of learning | No learning method |
| 5 | Combination of IT & Human Brain | Only deal with IT |
| 6 | Cannot be programmed | Can be programmed |

Henceforth from TABLE I it can be seen that ANN are better suited for the problem that are not so well defined and predictable. Further ANN's advantage is due to its clustering unlike other conventional systems .Hence ANN is betted suited for the problems that are not so well defined and predictable.

Applying ANN to HR factors graphically has been shown in fig 2.

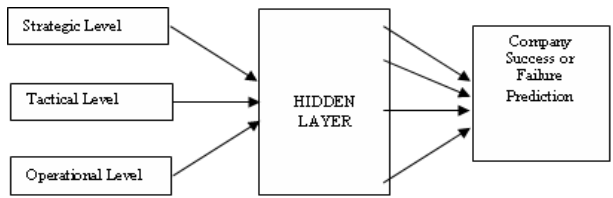

Fig. 2 Levels of HR of IS with ANN

## V RESEARCH METHODOLOGY

### A. Sampling scheme

The research involves the collection of data from the managers working at various levels within the selected enterprises. The total number of respondents in these enterprises, the sample size selection and application of the neural network approaches has been followed. The study comprises of survey of employees of two telecommunication companies. With this aim, two prestigious companies (first one is Reliance Communications, Chandigarh and the other one is Puncom, Mohali) have been considered in this study. The details of the research methodology adopted in this research are given below.

1) *For the Organisation:*
   a) Universe of study: Telecommunication industry comprises of Reliance InfoCom Vodafone, Essar, Idea, and Bharti-Airtel.
   b) Sample Selection: Reliance InfoCom, Chandigarh and Punjab Communication Ltd(Puncom) Mohali.

2) *For the Respondents*

a) Universe of study : All managers working at the three levels of the selected organizations.
b) Sample Selection: A number of respondents based on proportional stratified sampling from all of these organizations will be selected. The respondents will be identified from various levels in each organization. The sample size from a stratum was determined on the basis of the following criterion:

50% of the population where sample size > 5
100% of the population where sample size < 5.

### B. Data collection tools

Primary data has been collected through a questionnaire-cum-interview method from the selected respondents (Appendix C). The questionnaire was designed based on the literature survey, and detailed discussion with many academicians, professionals and industry experts. The detailed sampling plan of both the organizations has been shown in Table II.

TABLE II
DETAILS OF THE SAMPLING PLAN
PUNCOM, MOHALI AND RELIANCE, CHANDIGARH

| Level | Designation | Universe | Sample | % age | Total Sample |
|---|---|---|---|---|---|
| I | Executive Director (MD) | 2 | 2 | 100 | 17 |
|  | General Manager | 7 | 7 | 100 |  |
|  | Deputy General Manager | 6 | 8 | 50 |  |
| II | Assistant General Manager | 10 | 7 | 70 | 17 |
|  | Senior Manager | 10 | 5 | 50 |  |
|  | Manager | 10 | 5 | 50 |  |
| III | Deputy Manager | 30 | 15 | 50 | 135 |
|  | Senior Officer | 90 | 45 | 50 |  |
|  | Officer | 150 | 75 | 50 |  |

### C. Processing of data

The responses of the 169 managers of the selected organizations under study were recorded on five-point likert scale with scores ranging from 1 to 5. The mean scores of the managers of the two organizations and those over the whole industry considering all the managers included in the study.

The valid responses were entered in Microsoft Excel software. Thus, this data formed has been the basis for the corresponding files on the ANN software. The 70% responses of total inputs scores along with their known target from MS-Excel sheet were fed for training the neural network. The remaining scores of 30% responses were fed during the testing. Then the error_min of testing found to be less than the error_min of training data. The accuracy of 99.90% is shown in Table III.

## VI. EXPERIMENT RESULT AND DISCUSSION

### A. Dataset

The investigations have been carried out on the data obtained from telecommunication sector industry. This industry comprises of Reliance Communication, Vodafone, Essay, Idea, and Bharti-Airtel. But the data has been undertaken at Reliance Communication, Chandigarh and Punjab Communication Ltd. (Puncom) Mohali. The specific choice has been made because:

- The Telecom sector is very dynamic and fast growing. India is the second largest country of the world in mobile usage.
- The first industry is the early adopters of IT and has by now, gained a lot of growth and experience in IS development and whereas the other one lag behind and leads to its failure.

One industry is considered for the study because of the fact that the working constraints of various organizations under one industry are similar and hence adds to the reliability of the study finding. The input and output variables, considered in the study, include strategic parameter(x1), tactical parameter (x2), operational parameter (x3), employee outcome (y). The dataset comprises of 52 patterns has been considered for the training purpose of ANN and the remaining 23 patterns for testing the network.

### B. Connectionist model

The well versed 'trial and error' approach has been used throughout in this study. The Neural Network Toolbox under MATLAB R2007a is used for all training as well as simulation experiments.

1) Collected scores for both input data and known target were entered in MS-Excel as an input.
2) The input data of 70% of total data were imported to MAT lab's workspace for training the ANN as depicted in Appendix A.
3) The known target data has been also imported to Mat lab's workspace.
4) Then both the input data & the target data were entered in the ANN toolbox and network is created using back propagation neural network.
5) The training has been done using 70% of the input data and then testing (simulation) has been done on the rest of the 30% of the available data(Appendix B). The N/W used was backpropagtion with training function, traingda and adaptation learing function, learngdm. The mean square error MSE was found to be 0.096841. The accuracy of connectionist model for the prediction of success and/or failure of company results out to be 99.90%

Before analysis it is important to define:

*HL:* Hidden Layer (e.g. HL1: first Hidden Layer; HL2: second hidden Layer)

*Epoch:* During iterative training of Neural Network, an epoch is a single pass through the entire training set, Followed by the testing of the verification set.

*MSE:* Mean Square Error Learning Rate Coefficient $\eta$ -It determines the size of the weight adjustments made at each iteration which influence the rate of convergence.

The description of the Simulation Results of the Table III has been explained as

Col-1    It includes the configuration of the network having hidden layers 1(HL1) with 1 neuron and training function tansigmoidal, which remain same from 35-1000 epochs. Then 2/logsig tried for HL1 in the network, it has 2 neurons and HL2 i.e. hidden layer 2 having training faction tansigmodal tried for 35-1000 epoch. In this way the no. of neurons, training functions and hidden layers have been changed during trial and error method.

Col-2    No. of epoch (defined earlier) varies from 35-1000 per cycle

Col-3    Error goal is predefined for its tolerance.

Col-4    Learning Coefficient

Col-5    Mean Square Error for which the network is trained

Supervised feed-forward back propagation connectionist models based on viz., gradient descent algorithm has been used. The network was investigated empirically with a single hidden layer containing different numbers of hidden neurons and gradually more layers has added as depicted in the Table III. Several combinations of different parameters such as the data partitioning strategy; the number of epochs; the performance goal; transfer functions in hidden layers are explored on trial and error basis so as to reach the optimum combination.

The performance of the models developed in this study is evaluated in terms of mean square error (MSE) for the connectionist model using the neural tool kit. The mean square error indicates the accuracy for the prediction of success and/or failure of the organization comes out to be 99.90% through this model. The experimental results of simulation of data of success or Failure Company through this model are summarized in Table III.

TABLE-III
SIMULATION RESULTS (CONNECTIONIST MODEL)
USING ANN TOOLKIT

| Network Configuration | | Epochs | Error Goal | Learning rate | MSE |
|---|---|---|---|---|---|
| HL1 | HL2 | | | | |
|  |  | 35 |  |  | 0.738111 |
| 1/ tansig | - | 40 | 0.01 | 0.01 | 0.617595 |
| 1/ tansig | - | 45 | 0.01 | 0.01 | 0.634038 |
| 1/ tansig | - | 50 | 0.01 | 0.01 | 1.33051 |
| 1/ tansig | - | 80 | 0.01 | 0.01 | 0.580224 |
| 1/ tansig | - | 400 | 0.01 | 0.01 | 0.466721 |
| 1/ tansig | - | 1000 | 0.01 | 0.01 | 0.421348 |
| 2/ logsig | 1/ tansig | 35 | 0.01 | 0.01 | 1.22608 |
| 2/ logsig | 1/ tansig | 100 | 0.01 | 0.01 | 0.735229 |
| 2/ logsig | 1/ tansig | 200 | 0.01 | 0.01 | 0.402351 |
| 2/ logsig | 1/ tansig | 500 | 0.01 | 0.01 | 0.282909 |
| 2/ logsig | 1/ tansig | 1000 | 0.01 | 0.01 | 0.138904 |
| 3/ logsig | 1/ tansig | 35 | 0.01 | 0.01 | 0.81653 |
| 3/ logsig | 1/ tansig | 1000 | 0.01 | 0.01 | 0.143183 |
| 4/ logsig | 1/ tansig | 1000 | 0.01 | 0.01 | 0.096841 |

HL1: First hidden layer
HL2: Second hidden layer

The table-III shows when first hidden layer has 4 neurons and second hidden layer has 1 neuron with 1000 epochs, error goal 0.01, learning rate 0.01 mean square root is 0.096841, therefore accuracy of connectionist model for the prediction of failure company becomes 99.90%.

For further improvement Back propagation Approach has been deployed to reach better results. BPNN Code was written that generates error value for 1 to 2000 epochs and has shown the change in mean square error value.

C. *Back Propagation Algorithm*
For each input pattern do the following steps.
Step 1. Set parameters eata $\eta$ (............),
emax(maximum error value) and e(error between output and desired output).
Step2. Generate weights for hidden to output and input to hidden layers.
Step 3. Compute Input to Output nodes
Step4. Compute error between output and desired output
Step5. Modify weights from hidden to output and input to hidden nodes.

Result: This gives us error value of error=0.014289, no.of epochs=1405 during training It showed error_minima at epoc 1405 and its weights has been saved for feeding to testing algorithm.

The algorithm has been tested with 30% of data selected randomly from the given data which results in error=0.009174, no. of epochs=13 by doing programming of BPNN algorithium using Mat lab as shown in Table IV.

The results from the programming code have been shown through Matlab.

TABLE-IV
BPNN CODE TESTING RESULTS

```
`  MSE
   ↓
error=1.664782 no.of epoches=1
error=1.496816 no.of epoches=2
error=1.093136 no.of epoches=3
error=0.547380 no.of epoches=4
error=0.476718 no.of epoches=5
error=0.429089 no.of epoches=6
error=0.370989 no.of epoches=7
error=0.303513 no.of epoches=8
error=0.225575 no.of epoches=9
error=0.142591 no.of epoches=10
error=0.071286 no.of epoches=11
error=0.027775 no.of epoches=12
error=0.009174 no.of epoches=13
```

During testing the BPNN coding, error_minima has found to be less than error_minima of training, which validates the algorithm .It, has been further added that this accuracy of BPNN algorithium is found to be 99.99% whereas it was 99.90 in the connectionist model. Therefore this result is better than result obtained through hit and trail method (connectionist model) using neural network toolkit and hence BPNN algorithm's coding has fast performance and better results i.e.better prediction on low number of epochs at the time of testing could be achieved. During testing error_minima is less than error_minima of training for remaining 30% data, which validates algorithm. It comes out to be error=0.009174, at no. of epochs=13.Therefore the accuracy of the coding of the BPNN algorithium for the failure model comes out to be 99.99%.

VII CONCLUSION

HR factors have strong influence over company success and failure. Earlier HR factors were measured through variance estimation and statistical software's, Due to the inherent advantages of the Artificial Neural networks ,they are being used to replace the existing statistical models. Here, the ANN based model has been proposed that can be used for testing the success and/or failure of human factors in any of the communication Industry. Results have been obtained on the basis of connectionist model, which has been further refined by BPNN to an accuracy of 99.99%. Any company on the basis of this model can diagnose failure due to HR factors by directly deploying this model. The limitation of the study is that it only suggests a diagnostic model of success/failure of HR factors but it does not pin point them.

AUTHORS PROFILE

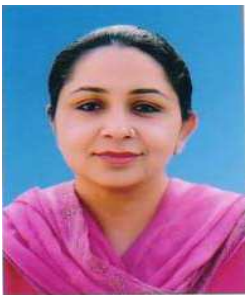

[1] **Bikram Pal Kaur** is an Assistant Professor in the Deptt. of Computer Science & Information Technology and is also heading the Deptt. Of Computer Application in Chandigarh Engineering College,Landran,Mohali. She holds the degrees of B.tech.,M.Tech,M.Phil.. and is currently pursuing her Ph.D.in the field of Information Systems from Punjabi University,Patiala. She has more than 11 years of teaching experience and served many academic institutions. She is an Active Researcher who has supervised many B.Tech.Projects and MCA Dissertations and also contributed 12 research papers in various national & international conferences. Her areas of interest are Information System, ERP.

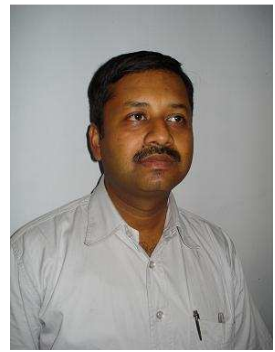

[2] **Dr. Himanshu Aggarwal**, is Associate Professor (Reader) in Computer Engineering at University College of Engineering, Punjabi University, Patiala. He had completed his Bachelor's degree in Computer Science from Punjabi University Patiala in 1993. He did his M.E. in Computer Science in 1999 from Thapar Institute of Engineering & Technology, Patiala. He had completed his Ph.D. in Computer Engineering from Punjabi University Patiala in 2007.He has more than 16 years of teaching experience. He is an active researcher who has supervised 15 M.Tech. Dissertations and guiding Ph.D. to seven scholars and has contributed more than 40 articles in International and National Conferences and 22 papers in research Journals. Dr. Aggarwal is member of many professional societies such as ICGST, IAENG. His areas of interest are Information Systems, ERP and Parallel Computing. He is on the review and editorial boards of several refereed Research Journals.

APPENDIX A

Table for **training data** is as following(70% data used for TRAINING)

|  | Strategic | Tactical | Operational |
|---|---|---|---|
| Emp1 | 1 | 2 | 1 |
| Emp2 | 2 | 3 | 1.9 |
| Emp3 | 4 | 1.5 | 1.5 |
| Emp4 | 2 | 3 | 4 |
| Emp5 | 1.7 | 1.6 | 2.5 |
| Emp6 | 1 | 1 | 1 |
| Emp7 | 1.2 | 1.3 | 1.4 |
| Emp8 | 1.7 | 1.8 | 3 |
| Emp9 | 1.8 | 2 | 4 |
| Emp10 | 4 | 1.8 | 2 |
| Emp11 | 2 | 5 | 1 |
| Emp12 | 2.5 | 2.2 | 2 |
| Emp13 | 2.5 | 2 | 1.6 |
| Emp14 | 1.6 | 2 | 2.5 |
| Emp15 | 1 | -1 | 1 |
| Emp16 | -1 | -1 | 1 |
| Emp17 | -1 | 1 | -1 |
| Emp18 | 1 | 1 | -1 |
| Emp19 | 1.2 | -1 | 1 |
| Emp20 | -1 | 1.2 | 1.5 |
| Emp21 | 3.6 | 1.2 | 4 |
| Emp22 | 3.6 | 3.6 | 3.6 |
| Emp23 | 4 | 4 | 5 |
| Emp24 | 5 | 4 | 2 |
| Emp25 | 5 | 5 | 1 |
| Emp26 | 4 | 5 | -1 |
| Emp27 | 3 | 2 | -1 |
| Emp28 | 0.5 | 1.5 | 0.5 |
| Emp29 | 2.1 | 3.1 | 4.1 |
| Emp30 | 5 | 5 | 5 |
| Emp31 | 0.1 | 0.2 | 0.5 |
| Emp32 | 0.5 | 0.7 | 1.5 |
| Emp33 | 4.1 | 4.2 | 4.3 |
| Emp34 | 5 | 0.1 | 0.2 |
| Emp35 | 0.1 | 2 | 0.1 |
| Emp36 | 1.4 | 5 | -1 |
| Emp37 | 1.5 | 4 | 1 |
| Emp38 | 1.6 | 3 | 2 |
| Emp39 | 2.1 | 2 | 3 |
| Emp40 | 2.1 | 1 | 4 |
| Emp41 | 2.3 | 5 | 5 |
| Emp42 | 2.5 | 4 | -1 |
| Emp43 | 3.3 | 3 | 1 |
| Emp44 | 3.5 | 2 | 2 |
| Emp45 | 4 | 1 | 3 |
| Emp46 | 4.9 | 5 | 4 |
| Emp47 | 4.1 | 4 | 5 |
| Emp48 | 4.3 | 3 | -1 |
| Emp49 | 3.01 | 2 | -1 |
| Emp50 | 2.01 | -1 | 1 |
| Emp51 | 2.03 | 5 | 1 |
| Emp52 | 5 | 4 | 1 |

## APPENDIX B

Table shows data used **for testing** neural network (30% data used for TESTING)

| | Strategic | Tactical | Operational |
|---|---|---|---|
| **Empt1** | **1.3** | **1.2** | **1.1** |
| Empt2 | 1.5 | 1.5 | 1.5 |
| Empt3 | 1.7 | 1.5 | 1.6 |
| Empt4 | 2 | 1 | 0 |
| Empt5 | 3 | 2 | 2 |
| Empt6 | 1.6 | 1.6 | 1.6 |
| Empt7 | 4 | 1 | 2 |
| Empt8 | 1 | 4 | 1.6 |
| Empt9 | 2 | 4 | 4 |
| Empt10 | 3.3 | 3.1 | 3.4 |
| Empt11 | 2.5 | 3.5 | 2 |
| Empt12 | 4.1 | 3.5 | 2.1 |
| Empt13 | 1 | 1 | 1 |
| Emptl4 | 1.3 | 1.1 | 1.9 |
| Emptl5 | 1.8 | 2.3 | 2.1 |
| Emptl6 | 0 | 0 | 0 |
| Emptl7 | 0 | 1 | 0 |
| Emptl8 | 0 | 0 | 1 |
| Emptl9 | 3.5 | 4.5 | 5 |
| Emptl20 | 1.8 | 1.6 | 2.9 |
| Emptl21 | 1 | 0 | 0 |
| Emptl22 | 2.5 | 1 | 1 |
| Emptl23 | 3.5 | 1.6 | 1.7 |

## APPENDIX -C

Questionnaire used for survey (containing scores from 1-5)
1-not important,2-slightly important,3-moderately important,4-fairly important,5-most important

| Factors | Score |
|---|---|
| **Strategic Factors** | |
| Support of the Top management | |
| Working relationship in a team(Users & Staff) | |
| Leadership | |
| project goals clearly defined to the team | |
| Thorough Understanding of business environment | |
| User involvement in development issues | |
| Attitude towards risk (Changes in the job profile due to the introduction of the computers) | |
| Adequacy of computer facility to meet functional requirements(quality and quantity both) | |
| Company technology focused | |
| Over commitment in the projects | |
| **Tactical Factors** | |
| Communication | |
| Organizational politics | |
| Priority of the organizational units to allocate resources to projects | |
| Organizational culture | |
| Skilled resources (Ease in the use of system by users) | |
| The consistency and reliability of information | |
| To obtain highest returns on investment through system usage | |
| Realization of user requirements | |
| Security of data and models from illegal users | |
| Documentation ( formal instructions for the usage of IS) | |
| The balance between cost and benefit of computer based information product/services | |
| User training | |
| The flexibility in system for corrective action in case of problematic output | |
| Testing of system before implementation | |
| **Operational factors** | |
| Professional standard maintenance (H/W, S/W, O.S, User Accounts, Maintenance of system) | |
| The response of staff to the changes in existing system | |
| Trust of staff in the change for the betterment of the system | |
| The way users input data and receive output | |
| The accuracy (Correctness) of the output | |
| The completeness (Comprehensiveness) of the information | |
| The well defined language for interaction with computers | |
| The volume of output generated by the system for a user | |
| Faith in technology/system by the user | |